\documentclass[nohyperref]{article}

\usepackage{mystyle}
\usepackage{annotate}

\usepackage{microtype}
\usepackage{graphicx}
\usepackage{subfigure}
\usepackage{booktabs}
\usepackage{hyperref}

\usepackage{amsmath}
\usepackage{amssymb}
\usepackage{mathtools}
\usepackage{amsthm}
\usepackage[inline]{enumitem}
\usepackage[dvipsnames]{xcolor}
\usepackage{multirow}
\usepackage{epstopdf}

\usepackage[accepted]{icml2022}

\newcommand{\myparagraph}[1]{\vspace{1mm}\noindent\textbf{#1} }

\theoremstyle{plain}
\newtheorem{theorem}{Theorem}[section]
\newtheorem{proposition}[theorem]{Proposition}

\newtheorem{remark}[theorem]{Remark}
\theoremstyle{definition}

\usepackage[capitalize,noabbrev]{cleveref}

\icmltitlerunning{Unsupervised Ground Metric Learning Using Wasserstein Singular Vectors}

\begin{document}

\twocolumn[
\icmltitle{Unsupervised Ground Metric Learning Using Wasserstein Singular Vectors}

\begin{icmlauthorlist}
    \icmlauthor{Geert-Jan Huizing}{dma,ibens}
    \icmlauthor{Laura Cantini}{ibens}
    \icmlauthor{Gabriel Peyré}{dma}
\end{icmlauthorlist}

\icmlaffiliation{dma}{D\'epartement de math\'ematiques et applications de l'Ecole Normale Sup\'erieure, CNRS, Ecole Normale Sup\'erieure, Universit\'e PSL, 75005, Paris, France}
\icmlaffiliation{ibens}{Computational Systems Biology Team, Institut de Biologie de l'Ecole Normale Sup\'erieure, CNRS, INSERM, Ecole Normale Sup\'erieure, Universit\'e PSL, 75005, Paris, France}

\icmlcorrespondingauthor{Geert-Jan Huizing}{huizing@ens.fr}
\icmlcorrespondingauthor{Gabriel Peyré}{gabriel.peyre@ens.fr}

\icmlkeywords{Machine Learning, ICML, Wasserstein, Optimal Transport, Metric Learning}

\vskip 0.3in
]

\printAffiliationsAndNotice{}  %

\newcommand{\A}{A}
\renewcommand{\a}{a}
\newcommand{\C}{\mathbb{A}}
\newcommand{\B}{B}
\renewcommand{\b}{b}
\renewcommand{\D}{\mathbb{B}}

\begin{abstract}
    Defining meaningful distances between samples in a dataset is a fundamental problem in machine learning.
    Optimal Transport (OT) lifts a distance between \textit{features} (the ``ground metric") to a geometrically meaningful distance between \textit{samples}.
    However, there is usually no straightforward choice of ground metric.
    Supervised ground metric learning approaches exist but require labeled data. In absence of labels, only ad-hoc ground metrics remain.
    Unsupervised ground metric learning is thus a fundamental problem to enable data-driven applications of OT.
    In this paper, we propose for the first time a canonical answer by simultaneously computing an OT distance between \textit{samples} and between \textit{features} of a dataset.
    These distance matrices emerge naturally as positive singular vectors of the function mapping ground metrics to OT distances.
    We provide criteria to ensure the existence and uniqueness of these singular vectors.
    We then introduce scalable computational methods to approximate them in high-dimensional settings, using stochastic approximation and entropic regularization.
    Finally, we showcase \textit{Wasserstein Singular Vectors} on a single-cell RNA-sequencing dataset.
\end{abstract}

\section{Introduction}
\label{sec-intro}

\begin{figure}[h!]
    \centering
    \vspace{1em}
    \includegraphics[width=.47\textwidth]{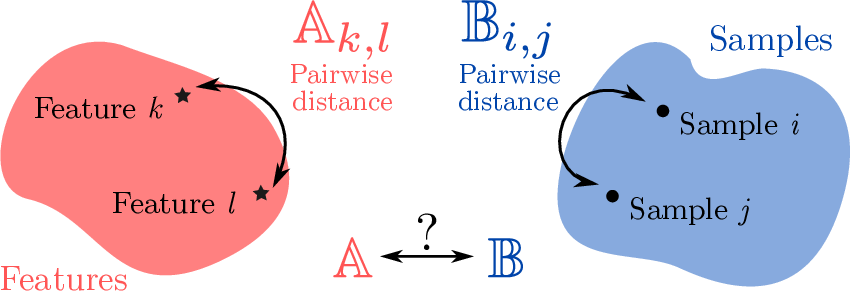}
    \caption{How to jointly define a distance $\C$ between features and a distance $\D$ between samples? Wasserstein Singular Vectors define natural Wasserstein distances $(\C,\D)$ in an unsupervised manner.}
    \label{fig:my_label}
\end{figure}

Machine learning tasks like information retrieval and classification require a notion of distance between samples in a dataset $X\in\RR^{n\x m}$. In particular, we will study the case of single-cell RNA-sequencing data (scRNA-seq).

\newpage
\myparagraph{Discrete histograms} In scRNA-seq data, the sample $x_i\in\RR_+^m$ represents the expression value of all genes in the $i$-th cell. This motivates two key assumptions in this paper: (i) samples are positive, which is natural when quantifying presence and quantity of physical objects (ii) samples can be normalized to discrete histograms, which is natural when the distribution over features (e.g. genes) is more important than the total mass. Indeed, gene expression in scRNA-seq data is usually normalized in some way as part of preprocessing~\cite{luecken2019current}.

\myparagraph{Optimal Transport distances}
Optimal Transport (OT)~\cite{Monge1781, Kantorovich42} offers a geometrically meaningful distance between discrete probability distributions, and has recently emerged as a useful tool for machine learning applications~\cite{frogner2015learning,pmlr-v51-rolet16}.
Contrarily to the Euclidean distance, OT does not compare distributions bin by bin.
Instead, OT lifts a ground pairwise distance matrix $\C\in\RR_+^{m\x m}$ between the $m$ features to the ``Wasserstein'' OT distance between normalized samples $a_i \eqdef x_i / \norm{x_i}_1$. It optimizes a transport plan $P\in\RR^{m\x m}$ encoding the displacement of mass between the two $m$-dimensional histograms $a_i, a_j$.
\begin{equation*}
    \Wass_\C(\a_i, \a_j) \eqdef 
    \!\!\min_{P\in\RR_+^{m\x m}} \textstyle\sum_\kl P_\kl \C_\kl ~\st~ 
    \begin{cases}
        P\ones_m = \a_i,\\P^\top\ones_m = \a_j.
    \end{cases}
\end{equation*}

\myparagraph{From supervised to unsupervised ground metric learning}
The crucial aspect of successful application of OT in ML is the design of a metric $\C$ which encodes the geometric relationships between features. 
In a supervised setting, one might take advantage of so-called \textit{ground metric learning} methods \cite{cuturi2014ground}. In an unsupervised setting, one usually resorts to some ad-hoc choice of ground cost. For instance the Word Mover Distance \cite{kusner2015word} uses Euclidean distances on Word2Vec embeddings \cite{mikolov2013efficient}. Similarly, the Gene Mover Distance \cite{bellazzi2021gene} uses Euclidean distances on Gene2Vec embeddings \cite{du2019gene2vec}.

In this work, we take a radically different route, by requiring that $\C_{k,\ell}$ is itself a Wasserstein distance between histograms $\b_k$ and $\b_\ell\in\RR_+^n$. The most intuitive case is to consider $a_i$ (resp. $b_k$) to be the normalized row $i$ (resp. column $k$) of a dataset $X\in\RR_+^{n\x m}$. We will thus refer to $a_i$ as a \textit{sample} and $b_k$ as a \textit{feature}.

An intuitive way to motivate our method is to consider a bootstrapping approach. 
Given some initial (for instance random) metric $\D$, one can compute $\C_{k,\ell} = \Wass_\D( b_k,b_\ell )$. But there is no reason to stop here, and the metric $\D$ can be ``improved'' by updating $\D_{i,j} = \Wass_\C( a_i,a_j )$. By continuing this process of successively updating $\C$ and $\D$, one could hope to reach a limit where the pair of distance matrices $(\C,\D)$ satisfies the following fixed point equation

\vspace{2em}

\begin{equation}\label{eq-sing-vec-intro}
    \tikzmarknode{AA}{\highlight{red}{$\C_\kl$}} = \ufrac{\lambda}\Wass_{\tikzmarknode{BB2}{\tighthighlight{blue}{$\D$}}}(\tikzmarknode{bb}{\highlight{red}{$\b_k, \b_\ell$}})
    ,~
    \tikzmarknode{BB}{\highlight{blue}{$\D_\ij$}} = \ufrac{\mu}\Wass_{\tikzmarknode{AA2}{\tighthighlight{red}{$\C$}}}(\tikzmarknode{aa}{\highlight{blue}{$\a_i, \a_j$}}),
\end{equation}
\begin{tikzpicture}[overlay,remember picture,>=stealth,nodes={align=left,inner ysep=1pt},<-]
      
      \path (AA.south) ++ (0,-1.8em) node[anchor=south west,color=red!60] (scalep){\strut Distance between features};
      \draw [color=red!60](AA.south) |- ([xshift=-0.3ex,color=red]scalep.south east);
      
      \path (aa.south) ++ (0,-1.8em) node[anchor=south west,color=blue!60] (scalep){\strut Samples};
      \draw [color=blue!60](aa.south) |- ([xshift=-0.3ex,color=red]scalep.south east);
      
      \path (bb.north) ++ (0,1.2em) node[anchor=south east,color=red!60] (scalep){\strut Features};
      \draw [color=red!60](bb.north) |- ([xshift=-0.3ex,color=red]scalep.south west);
      
      \path (BB.north) ++ (0,1.2em) node[anchor=south west,color=blue!60] (scalep){\strut Distance between samples};
      \draw [color=blue!60](BB.north) |- ([xshift=-0.3ex,color=blue]scalep.south east);
    \end{tikzpicture}
    
\vspace{1em}

where $(\lambda, \mu)\in\RR_+^2$ are scaling factors. This corresponds to casting ground metric learning as a non-linear singular vectors problem.
In this work, we study theoretical properties (in particular existence and uniqueness) as well as the practical relevance of Wasserstein Singular Vectors for machine learning.

\subsection{Previous works}

\myparagraph{Optimal Transport}
While the initial proposal of Monge~\cite{Monge1781} formulates the OT problem as an optimal matching problem, its modern and tractable formulation by Kantorovich~\cite{Kantorovich42} is a linear program detailed in the Introduction.
Besides its use to define matchings and couplings between distributions, the main feature of OT is that the transportation value induces a geometric distance on the space of probability distributions.
This ``Wasserstein'' distance is thus parameterized by the underlying ground cost between pairs of points.
We refer to the monographs \cite{villani2003,santambrogio2015optimal} for a detailed account of the theory of OT, and \cite{peyre2019computational} for its computational aspects.
OT distances have been used for applications as diverse as image retrieval~\cite{rubner-2000}, brain imaging  \cite{gramfort2015fast, janati2020spatio}, natural language processing \cite{kusner2015word,yurochkin2019hierarchical}, and generative models \cite{arjovsky2017wasserstein,tolstikhin2017wasserstein}. In recent years, many applications of OT to single-cell biology have been proposed \cite{hashimoto2016learning,schiebinger2019optimal,bellazzi2021gene,huizing2021optimal,tong2021embedding}.

\myparagraph{Entropic regularization}
Entropic regularization of OT allows to scale to high-dimensional machine learning problems. It approximates OT distances using Sinkhorn's algorithm, which has quadratic complexity and streams well on GPU architectures.
Entropic regularization was put forward in the seminal paper by Cuturi \cite{CuturiSinkhorn}, who also emphasizes the smoothing effect, which is crucial when using Sinkhorn as a loss function to train deep learning models.
Another benefit of this regularization is that it suffers less from the curse of dimensionality, as proved in~\cite{genevay2019sample, mena2019statistical}.
This approach is also pivotal to scale our unsupervised metric learning method to tackle high-dimensional problems, for instance in genomics.

\myparagraph{Metric Learning}
Metric learning is most often framed as the supervised problem of minimizing (resp. maximizing) the distance between points in a same (resp. different) class. Existing approaches are reviewed in \cite{kulis2012metric,bellet2013survey}.
It is necessary to restrict the class of distances to make the problem tractable. The most common option is arguably to consider the class of Mahalanobis distances, which generalize the Euclidean distance and are equivalent to computing a vectorial embedding of the data points. See for instance~\cite{xing2002distance,weinberger2006distance, davis2008structured}.
One can apply these methods for histogram data, or use instead of Euclidean distances more adapted discrepancies on the simplex, such as Chi-squared~\cite{noh2012chi,yang2015chi} and geodesic distances~\cite{le2015unsupervised}.
These methods however fail to capture the geometric nature of the problem, where histograms correspond to discrete distributions viewed as sums of localized Dirac masses.

\myparagraph{OT Ground Metric Learning}
This geometry is leveraged in \cite{cuturi2014ground} by introducing the problem of supervised OT ground metric learning and developing a nearest-neighbor based algorithm to solve it.
This approach is further refined in \cite{wang2012supervised}, which drops the triangular inequality constraint (as we do in our approach).
It is possible to restrict the class of ground metrics, for instance using Mahalanobis~\cite{xu2018multi, kerdoncuff:ujm-02611800} or geodesic distances~\cite{heitz2020ground} to develop more efficient learning schemes.
\cite{zen2014simultaneous} simultaneously perform ground metric learning and matrix factorization, and this finds applications to NLP~ \cite{huang2016supervised}.
Metric learning can also be performed through adversarial optimization, where the metric is maximized over to perform generative model training~\cite{2018-Genevay-aistats},  discriminant analysis~\cite{flamary2018wasserstein} and to define robust transportation distances~\cite{paty2019subspace,niles2019estimation}.
Note that when imposing only convex constraints, adversarial ground metric learning is a concave maximization problem which finds applications in the modeling of crowd congestion phenomena~\cite{benmansour2010derivatives}.
Another related question is the inverse problem of estimating a ground cost from the observation of matchings or couplings~\cite{galichon2020cupid,stuart2020inverse,li2019learning,paty2020regularized}. This supervised metric learning problem can be regularized using sparsity or low-rank constraints, as explained in~\cite{dupuy2019estimating,carlier2020sista}.
Finally, ``hierarchical OT''~\cite{yurochkin2019hierarchical, abrishami2020geometry} uses OT to define the ground cost of a matching problem, using an intermediate level of meta-features.

\subsection{Contributions}

Our main contribution is the introduction in Section~\ref{sec-wasserstein-singular-vectors} of \textit{Wasserstein singular vectors} as the positive singular vectors of monotone homogeneous ``distance maps''.
The associated theoretical contributions, Theorem~\ref{thm-existence} and Theorem~\ref{thm-uniqueness}, state conditions ensuring the existence and uniqueness of such singular vectors.

Our second set of contributions allows to scale the method to large datasets.
We first introduce in Section~\ref{sec-stochastic} a stochastic algorithm similar in spirit to Projected Stochastic Gradient Descent. Theorem~\ref{thm-stochastic} guarantees a convergence rate of $\Oo(\log t/\sqrt{t})$ under certain conditions.
We then explain in Section~\ref{sec-entropic} how to scale and parallelize this method even further by leveraging entropic regularization through the Sinkhorn algorithm.
Proposition~\ref{prop-pca} shows that in the large regularization limit, our method computes metrics associated to 1-D and 2-D embeddings along the leading principal component axes.

Section~\ref{sec-genomics} demonstrates the potential of Wasserstein Singular Vectors compared to ad-hoc applications of Optimal Transport, by studying a single-cell RNA-sequencing dataset.

A Python package implementing all algorithms in this paper is available at \href{https://github.com/gjhuizing/wsingular}{github.com/gjhuizing/wsingular}. Optimal Transport distances were computed using the open-source POT library \cite{flamary2017pot}. Appendix~\ref{supp-resources} lists the experiments' computation times and resources.

\subsection*{Notations}
We denote $\Dd_m \subset\RR_+^{m\x m} $ the set of pairwise distance matrices. In other words, $\C \in \Dd_m$ if
(i) $\C_\kl = 0\iff k = \ell$, 
(ii) $\C_\kl \leq \C_{k,s} + \C_{s,\ell}$ 
(iii) $\C_\kl = \C_{\ell,k}$. 
Its closure $\bar\Dd_m \eqdef \{ \C \in \RR_+^{m\x m} \st \C = \C^\top, \diag(\C) = 0\}$ is a set of pseudo-distances.

\section{Unsupervised Wasserstein Metric Learning}
\label{sec-wasserstein-singular-vectors}

This section introduces the singular vectors of the Wasserstein distance map. Fortunately, this non-linear singular vector problem enjoys many desirable properties.

\subsection{Wasserstein Singular Vectors} 

\myparagraph{Wasserstein distance map} The following map lifts a ground metric $\C \in \Dd_m$ to a pairwise distance matrix $\Phi_\A(\C) \in \Dd_n$. A norm $R$ operates as a regularizer to enforce strict positivity of the computed distances.

\begin{equation}
\label{def-wasserstein-distance-map}
    \Phi_\A(\C)_\ij \eqdef \Wass_\C(\a_i, \a_j) + \tau \norm{\C}_\infty R(\a_i - \a_j)
\end{equation}
The map $\D\in\Dd_n \mapsto \Phi_\B(\D) \in \Dd_m$ is defined similarly.

\myparagraph{Role of regularization $\mathbf\tau$~} Let us insist that in practice our method can be applied in the unregularized setting $\tau=0$, but some of the theoretical claims require $\tau>0$. Other types of regularization could be considered, for instance a matrix with zeros on the diagonal and ones elsewhere.

\myparagraph{Wasserstein singular vectors} With this notation, our ground metric learning solves for a pair $\C\in\Dd_m$ and $\D\in\Dd_n$ of \textit{Wasserstein singular vectors} satisfying 
\begin{equation}\label{eq-wass-sing-problem}
    \exists(\lambda, \mu)\in (\RR_+^*)^2
    \st
    \Phi_\B(\D)=\lambda \C, 
    \;
    \Phi_\A(\C) = \mu \D, 
\end{equation}
which corresponds to~\eqref{eq-sing-vec-intro} when $\tau=0$.
The case $m=n$ and $\A=\B$ corresponds to the computation of an eigenvector $\C=\D$ of $\Phi_\A$ with eigenvalue $\lambda=\mu$.

\myparagraph{Power iterations algorithm} The de-facto standard algorithm to extract singular vectors are ``power iterations''
\begin{align}\label{eq:poweriter}
    \C_{t+1} &\eqdef \frac{\Phi_\B(\D_{t})}{\norm{\Phi_\B(\D_{t})}_\infty}, 
    \quad
    \D_{t+1} \eqdef \frac{\Phi_\A(\C_{t+1})}{\norm{\Phi_\A(\C_{t+1})}_\infty}.
\end{align}
The complexity of performing a single power iteration is $\mathcal{O}(n^2 m^2 (n\log(n)+m\log(m)))$, since  the computation of a single Wasserstein distance in $\RR_+^n$ is  $\mathcal{O}(n^3\log n)$ \cite{bonneel2011displacement}. 
To cope with large scale datasets, we propose in Section~\ref{sec-stochastic} and \ref{sec-entropic} to use stochastic optimization and entropic regularization. 
\begin{remark}
A remarkable property of this algorithm is that, in cases where the singular vector is unique (which is observed in practice and proved below for large $\tau$), even if the initialization $\D_{t=0}$ is chosen arbitrarily, it converges toward a distance matrix (so in particular it satisfies the triangular inequality at convergence).
\end{remark}

\subsection{Theoretical Properties}
\label{subsec-wasserstein-singular-vectors}

\myparagraph{Properties of the Wasserstein distance map} Non-linear singular vectors problems are notoriously difficult to study. Fortunately, as explained in the Proposition~\ref{prop-wasserstein-distance-map}, the Wasserstein distance map is a so-called topical mapping (i.e. positive and monotone)~\cite{lemmens2012nonlinear}.
These mappings can be thought as non-linear generalizations of Markov chains. 
Problem~\eqref{eq-wass-sing-problem} is thus an instance of a non-linear Perron-Frobenius problem, for which, contrarily to generic problems, existence and uniqueness of positive solutions is in general rather the rule than the exception. This explains in large part the practical success of our approach.

The following proposition lists some useful properties of the Wasserstein distance map $\Phi_\A$.
\begin{proposition}
\label{prop-wasserstein-distance-map}
    \begin{enumerate*}[label=~(\roman*)~]
        \item $\Phi_\A$ is positively 1-homogeneous and monotone.
        \item $\Phi_\A$ is continuous on $\Dd_m$ %
        \item $\Phi_\A$ is $(1 + 2\tau k_R)$-Lipschitz on $\Dd_m$ with regards to the $\norm{\cdot}_\infty$ norm, where the constant $k_R>0$ is such that $R(\cdot)\leq k_R\norm{\cdot}_1$.
    \end{enumerate*}
\end{proposition}

\begin{proof}
(i) 1-homogeneity and monotony of $\Phi_\A$ follows from the definition.
(ii) Note that $\Phi_\A$ is a vector-valued concave function (each coordinate being an infimum of linear forms) and hence is continuous on $\RR^{m\x m}$. Actually, as we now show, it is Lipschitz for $\ell^\infty$.
(iii) Let us prove that $\Phi_\A$ is Lipschitz continuous for the $\ell^\infty$ norm on $\Dd_m$. Firstly, since $R(a_i - a_j) \leq k_R\norm{\a_i - \a_j}_1$ and $\left|\norm{\C}_\infty - \norm{\C'}_\infty\right|\leq\norm{\C - \C'}_\infty$, we have
\begin{align*}
    |\Phi_\A(\C)_\ij - \Phi_\A(\C')_\ij| \leq& \\
    |\Wass_\C(\a_i, \a_j) - \Wass_{\C'}(\a_i, \a_j)| + 2\tau k_R&\norm{\C - \C'}_\infty
\end{align*}
Secondly, with $\Gamma(\a, \a')$ the set of valid couplings,
\begin{align*}
    |\Wass_\C(\a&, \a') - \Wass_{\C'}(\a, \a')| \\
    &=|\min_{P\in\Gamma(\a, \a')}\dotp{P}{\C} - \min_{P'\in\Gamma(\a, \a')}\dotp{P'}{\C'}|\\
    &\leq \norm{\C - \C'}_\infty.
\end{align*}
Indeed, $|\min(u) - \min(v)|\leq\max |u - v|$ and $\norm{P}_1 = 1$. So $\Phi_\A$ is $(1 + 2\tau k_R)$-Lipschitz.
\end{proof}

\myparagraph{Existence of singular vectors} The following proposition ensures the existence of positive (i.e. true distances) singular vectors. Its proof can be found in Appendix~\ref{proof-existence}. Note that the $\ell^\infty$ bound of Proposition~\ref{prop-wasserstein-distance-map} implies that all singular values are smaller than $1+2\tau k_R$.~

\begin{theorem}
\label{thm-existence}
When $\tau>0$, there exist positive singular vectors $(\C, \D)\in\Dd_m\x\Dd_n$ solving the problem~\eqref{eq-wass-sing-problem}.
\end{theorem}

\myparagraph{Existence in the case $\tau=0$~}Extending Theorem~\ref{thm-existence} to the unregularized case is an open problem, and is out-of-reach using classical non-linear Perron-Frobenius theorems such as~\cite{akian2016uniqueness,akian2018game}, which do not apply. The following remark exhibits solutions for a special case of the unregularized problem.

\begin{remark}[Block-diagonal matrices]
Let us consider the case $\tau=0$ and a dataset $X=\diag(X_p)$, a block-diagonal matrix where $X_p \in \RR_+^{n_p \times m_p}$. Let $\A$ and $\B$ be its normalizations along rows and columns respectively. Then all matrices $(\C, \D)\in\bar \Dd_m\x\bar\Dd_n$ of the form $\C=(c_{p,q} \ones_{m_p \times m_q})_{p,q}$ and $\D=(c_{p,q} \ones_{n_p \times n_q})_{p,q}$ for the same $c_{p\neq q} \in \RR_+^*$ and $c_{p,p}=0$ are dominant singular vectors with associated singular value $1$. Indeed, the optimal transport plans for these ground costs also follow this block structure.
\end{remark}

\myparagraph{Uniqueness of singular vectors} 
If the dataset is too sparse and $\tau=0$, one cannot hope to have uniqueness of the leading singular vectors. In fact, the previous remark exhibits an infinity of singular vectors when $X$ is block-diagonal.
It does not seem obvious to guarantee uniqueness by an a priori condition depending only on $(A,B)$. The following proposition gives an a posteriori way to check the uniqueness of singular vectors inside $\Dd_m\x\Dd_n$. In the numerical simulations of Section~\ref{numerical-translated}, we checked \textit{a posteriori} that the computed singular vectors are indeed unique.

\begin{theorem}
\label{thm-uniqueness}
	Let $(\C, \D) \in \Dd_m\x\Dd_n$ a pair of Wasserstein Singular Vectors. 
	We consider $P(\a_i,\a_j)$ and $P(\b_k,\b_\ell)$ optimal coupling solutions of the OT problems for the costs $\C$ and $\D$ respectively.
	These optimal couplings induce a graph on $\{(i,j),(k,\ell)\}$ by linking
	\begin{align*}
	    (k,\ell) \rightarrow (i,j)&\text{ when }P(\a_i,\a_j)_\kl >0\\
	    (i,j) \rightarrow (k,\ell)&\text{ when }P(\b_k, \b_\ell)_\ij >0.
	\end{align*}
	If there exist optimal couplings such that this graph is strongly connected, then $(\C, \D)$ are the unique  Wasserstein singular vectors.
\end{theorem}

\begin{proof}Let us consider $\Phi : (\C,\D) \mapsto (\Phi_\B(\D), \Phi_\A(\C))$ which maps $\Dd_m \times \Dd_m$ to itself. We use Theorem~7.5 of~\cite{akian2016uniqueness}. It requires that the semi-differential of $\Phi$ at $(\C, \D)$ has itself a unique positive eigenvector in $\Dd_m\x\Dd_n$. In fact, this eigenvector is also $(\C, \D)$.

From the envelope theorem, upper-gradients of the concave function $\C \mapsto \Wass_\C(\a_i,\a_j)$ are the elements $P(\a_i,\a_j)$ (so if the optimal coupling is unique, then this map is differentiable). The semi-differential $\Psi$ of $\Phi$ is then a block anti-diagonal matrix defining the graph detailed in the statement of the theorem.
The (linear) Perron-Frobenius theorem for positive linear operators ensures the existence of a unique positive eigenvector of $\Psi$ if this graph is connected.
\end{proof}

\begin{figure*}[h]
    \centering
    \includegraphics[width=.96\textwidth]{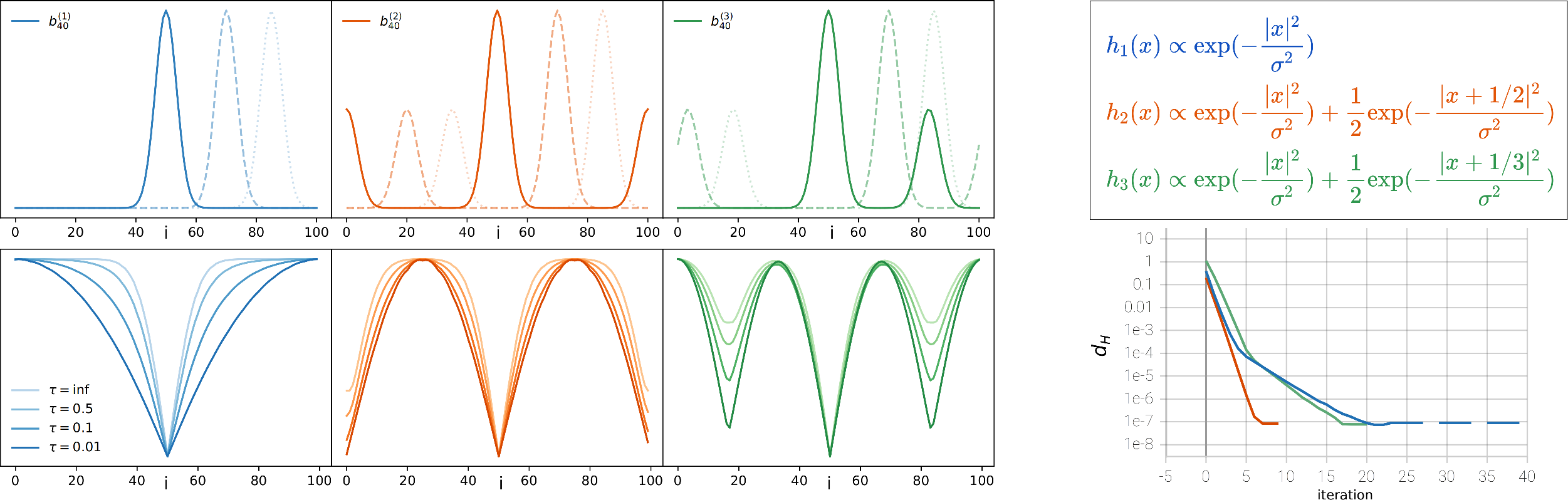}
    \caption{Illustration on the 1-D torus.
    (top, left) histograms whose translations form {\color{RoyalBlue} $\B_1$}, {\color{orange} $\B_2$}, {\color{OliveGreen} $\B_3$} ;
    (bottom, left) distance maps {\color{RoyalBlue} $c_1$}, {\color{orange} $c_2$}, {\color{OliveGreen} $c_3$} associated to the singular vectors {\color{RoyalBlue} ${\D_1}$}, {\color{orange} ${\D_2}$}, {\color{OliveGreen} ${\D_3}$} for varying values of $\tau$ ;
    (top, right) functions {\color{RoyalBlue} $h_1$}, {\color{orange} $h_2$}, {\color{OliveGreen} $h_3$} generating the datasets ;
    (bottom, right) convergence rate of the power iterations for $\tau=0.1$, according to the $d_\Hh$ metric.}
    \label{fig:num_1D_conv}
\end{figure*}

\myparagraph{Convergence of power iterations} 
In the case of linear positive maps, Perron-Frobenius theory ensures the convergence of~\eqref{eq:poweriter} toward the unique positive singular vectors at a linear rate.
Unfortunately, this result does not hold in general for the case of non-linear maps, and $\Phi_\A$ is only non-expansive (and not necessarily contracting). The following proposition, proved in Appendix~\ref{unique-large-tau}, states that for large enough regularization, uniqueness and linear convergence are maintained.
\begin{proposition}\label{prop:contractlarge}
    For $\tau$ large enough, the singular vectors are unique and the power iterations~\eqref{eq:poweriter} converge linearly for $\norm{\cdot}_\infty$. When $\tau \rightarrow \infty$, the singular vectors converge to  $\C_{k,\ell} \propto R(\b_k - \b_\ell)$ and $\D_{i,j} \propto R(\a_i - \a_j)$.
\end{proposition}
The numerical simulations of Section~\ref{numerical-translated} suggest that uniqueness and linear rates always hold in practical cases. 

\subsection{Numerical illustration on translated histograms}
\label{numerical-translated}

\myparagraph{Generating translated histograms} We generate three synthetic datasets
{\color{RoyalBlue} $X_1$}, {\color{orange} $X_2$}, {\color{OliveGreen} $X_3$}
$\in\RR^{n\times m}$ by translating three different templates.
We define the datasets by $[X_p]_{i,k} \eqdef h_p(i/n - k/m)$, and $\A_p$ (resp. $\B_p$) is obtained by normalizing $X_p$ along rows (resp. columns). By translational invariance of the problem, the singular vectors {\color{RoyalBlue} $\C_1$}, {\color{orange} $\C_2$} and {\color{OliveGreen} $\C_3$} are of the form $(\C_p)_{k,l}=(c_p)_{k-l}$ where $c_p=(\C_p)_{0,\cdot}$ are periodic 1-D functions. The same argument applies to the singular vectors {\color{RoyalBlue} $\D_1$}, {\color{orange} $\D_2$} and {\color{OliveGreen} $\D_3$}. The templates {\color{RoyalBlue} $h_1$}, {\color{orange} $h_2$}, {\color{OliveGreen} $h_3$} are three different periodic functions on the 1-D torus (we use periodic boundary conditions). We use $n=100$ samples and $m=80$ features.

\myparagraph{Wasserstein singular vectors} We compute the Wasserstein singular vectors for different values of  $\tau$. Figure \ref{fig:num_1D_conv} displays the templates and the corresponding singular vectors {\color{RoyalBlue} $\C_1$}, {\color{orange} $\C_2$} and {\color{OliveGreen} $\C_3$} obtained through power iterations. {\color{RoyalBlue} $\D_1$}, {\color{orange} $\D_2$} and {\color{OliveGreen} $\D_3$} are symmetric and can be found in Appendix~\ref{supp-numeric-translated}. These results demonstrate that the learned metrics integrate geometrical properties (symmetries, multi-modalities, etc.) of the input datasets. For unimodal Gaussian-like distributions, the learned metric is close to $|\sin(i/n-j/n)|$, but exhibits non-monotonic behavior for multi-modal distributions.

\myparagraph{Convergence rates} Figure \ref{fig:num_1D_conv} also reports in logarithmic scale the convergence rate of power iterations according to the Hilbert metric 
$d_\Hh(\D, \D') \eqdef \norm{\log(\D/\D')}_V$ where $\norm{Z}_V \eqdef \max(Z)-\min(Z)$.
This speed is always linear, suggesting that the maps $\Phi_{\A_p}$ are contracting and that the singular vectors are unique (which is confirmed by running several initializations in $\Dd_n$, and through condition~\ref{thm-uniqueness}).
The contractance rate (which is the slope of the error curves) is dependent on the geometry of the templates~$h_p$. We also observed a steeper slope for larger values of $\tau$.

\section{Large Scale Stochastic Power Iterations}
\label{sec-stochastic}

As $n$ or $m$ grows, the complexity of the power iterations~\eqref{eq:poweriter} becomes prohibitive. In order to work around this issue we propose a stochastic power iteration scheme similar in spirit to stochastic gradient descent, which updates a single (or several if applied in a mini-batch setting) randomly chosen distance value at each step.
This speeds up each iteration and leverages the correlations in the dataset.

\myparagraph{Stochastic power iterations} For some decreasing step size $\alpha_{t}$ and a scaling factors $\tilde\lambda_t,\tilde\mu_t > 0$, we define
\begin{align*}
    \C_{t+1} \eqdef \Pi( (1-\alpha_{t})\C_{t} + \alpha_{t}\tilde\C_{t} ), \\
    \D_{t+1} \eqdef \Pi(  (1-\alpha_{t})\D_{t} + \alpha_{t}\tilde\D_{t} ),
\end{align*}
$$ \mathrm{where}\quad (\tilde\D_{t})_\ij \eqdef \begin{cases}
        \Phi_\A(\C_{t})_\ij/{\tilde\mu_t} \text{ if }(\ij)=(i_t,j_t),\\
        (\D_{t})_\ij\text{ otherwise.}
    \end{cases}$$
We define $\Pi(\C) \eqdef \C/\norm{\C}_\infty$ and $(i_t,j_t)$ is is drawn uniformly at random in $\{1,\ldots,n\}^2$. 
It is the index updated at each step. $\tilde\C_t$ is computed by an analogous update rule, with $\tilde\mu_t$ replaced by $\tilde\lambda_t$.

\myparagraph{Convergence of stochastic power iterations} The following theorem, proved in Appendix~\ref{proof-thm-stochastic}, guarantees that for a large enough regularization parameter $\tau$, these iterations converge to a pair of Wasserstein Singular Vectors. In practice, we observe that these iterations converge even for arbitrary small $\tau$ and for $\tilde\lambda_t=\tilde\mu_t=1$.

\begin{theorem}%
\label{thm-stochastic}

For
$\alpha_{t} = 1/\sqrt{t}$,
for constant scaling factors
$\tilde\lambda_t \leq \tau\min_{k\neq l}R(b_k - b_l)$ and
$\tilde\mu_t \leq \tau\min_{i\neq j}R(a_i - a_j)$,
and for $\tau$ large enough,
the stochastic power iterations defined above converge to a pair $(\C, \D)\in\Dd_m\x\Dd_n$ of positive singular vectors with a convergence rate of $\Oo(\log(t)/\sqrt t)$.
\end{theorem}

\begin{remark}[Adaptive selection of $\tilde\lambda_t$ and $\tilde \mu_t$]
\label{remark-estimation}
Tuning the values of the parameters $\tilde\lambda_t$ and $\tilde \mu_t$ is crucial to improving the convergence. Ideally, they should be as close as possible to the (unknown) singular values $(\lambda,\mu)$.   
Instead of fixing these scaling factors prior to running the algorithm, we propose  using an estimation of the singular values. When using mini-batching, i.e. updating several indices $(i, j)\in\mathcal I$ at each iteration, one can use a least square estimate,
 $$\tilde \mu_{t+1} = (1 - \alpha_t)\tilde\mu_t + \alpha_t\frac{\sum_{\ij\in\mathcal I}\Phi_A(\C_t)_\ij(\D_t)_\ij}{\sum_{\ij\in\mathcal I}(\D_t)_\ij^2},$$
 and similarly for $\tilde \lambda_{t+1}$. We found that in practice, $(\tilde \lambda_t,\tilde \mu_t)$  quickly converges to the singular values $(\lambda,\mu)$.
\end{remark}

\begin{figure}[h]
    \centering
    \includegraphics[width=.38\textwidth]{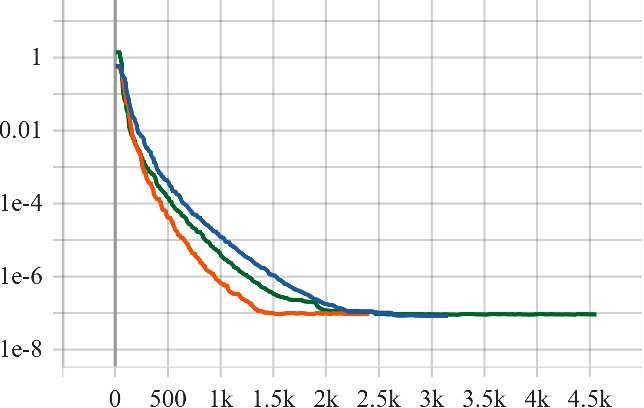}
    \caption{Convergence rates $d_\Hh(\D_t, \D_\infty)$ of stochastic updates.}
    \label{fig:stochastic}
\end{figure}

\myparagraph{Numerical illustration on translated histograms} Figure~\ref{fig:stochastic} illustrates the convergence of stochastic power iterations in practice by comparing the approximated Wasserstein Singular Vectors for the synthetic experiments of Section~\ref{numerical-translated} with the true singular vectors obtained using the non-stochastic power iterations~\eqref{eq:poweriter}. We used the approach outlined in Remark~\ref{remark-estimation}. As expected, the stochastic power iterations yield the same result as classical power iterations.

\section{Parallelization With Entropic Regularisation}
\label{sec-entropic}

To further speed up the method, we propose to use the entropic regularization of Optimal Transport~\cite{CuturiSinkhorn}.

\myparagraph{Sinkhorn's algorithm} Entropic OT can be computed efficiently in $\mathcal O(n^2/\eta^2)$ using Sinkhorn's algorithm (detailed in Appendix~\ref{supp-sinkhorn}) at the expense of an approximation of order $\eta$~\cite{altschuler2017near}. 
Beside speeding up the computation of OT, this enables embarrassingly parallel computations of the distance map on GPUs~\cite{CuturiSinkhorn} and also reduces the curse of dimensionality which plagues OT~\cite{genevay2018sample}.

\begin{remark}
The theoretical guarantees listed above apply to entropic OT with Euclidean ground costs, but not necessarily to the setting described in this paper. However, in practice we observed that the benefits of Sinkhorn's algorithm were maintained in our situation.
\end{remark}

\subsection{Sinkhorn divergence map}

\myparagraph{Sinkhorn divergence} Entropic regularized OT is defined
$$
    \Wass_\C^\eps(\a_i, \a_j) \eqdef \min_{P\in\Gamma(\a_i, \a_j)} \dotp{P}{\C} + \eps\norm{\C}_\infty \dotp{P}{\log P}.
$$
This quantity suffers from a bias, which is removed by using instead the Sinkhorn divergence~\cite{2018-Genevay-aistats} 
$$
    \bar \Wass_\C^\eps(\a_i, \a_j) 
    \!\eqdef\! 
    \Wass_\C^\eps(\a_i, \a_j) - \ufrac{2}\left(\Wass_\C^\eps(\a_i, \a_i) \!+\! \Wass_\C^\eps(\a_j, \a_j)\right).
$$
This debiasing is crucial to ensure that $\bar \Wass_{\C}^\eps(\a_i,\a_i)=0$, and it also reduces the approximation error to $|\bar\Wass_{\C}^\eps-\bar\Wass_{\C}|\sim \eps^2$~\cite{2020-Chizat-EntropicFaster}.

\myparagraph{Sinkhorn divergence map} Similarly to the map defined in~\eqref{def-wasserstein-distance-map}, we define the Sinkhorn divergence map as 
$$
    \Phi_\A^\eps(\C\in\Dd_m)_\ij \eqdef \bar \Wass_\C^\eps(\a_i, \a_j) + \tau \norm{\C}_\infty R(\a_i - \a_j), 
$$
and similarly for $\Phi_\B^\eps$. It reduces to~\eqref{def-wasserstein-distance-map} when $\eps=0$.  

\myparagraph{Sinkhorn singular vectors} By analogy with \eqref{eq-wass-sing-problem} we define \textit{Sinkhorn singular vectors} as $\C, \D$ such that
\begin{equation}\label{eq-sink-sing-problem}
    \exists(\lambda, \mu)\in{\RR_+^*}^2
    \st
    \Phi^\eps_\B(\D)=\lambda \C, 
    \;
    \Phi^\eps_\A(\C) = \mu \D, 
\end{equation}

\begin{remark}[Positivity property] 
The map $\Phi^\eps$ is 1-homogeneous and monotone, but it is unclear that it always maps onto positive matrices. 
It is proved in \cite{feydy2019interpolating} that it is the case if $e^{-\C/(\eps \norm{\C}_\infty)}$ is a positive kernel (i.e. has positive eigenvalues). While it is unclear that such a condition is maintained during power iterations, we observed numerically that it is still the case in practice.
We show below that this is true in the limit $\eps\to +\infty$. 
\end{remark}

\subsection{Connection with PCA when $\eps\to\infty$}

\myparagraph{Maximum Mean Discrepancy limit} For simplicity, let us consider the case $\tau=0$. We show in  proposition~\ref{prop:mmd} below that when $\eps\to\infty$, our method operates over the set of squared Euclidean matrices
$$
    \C \in \Kk_m \subset \Dd_m \iff\exists(u_k \in\RR^d)_{k=1}^m, \C_\kl = \norm{u_k - u_\ell}_2^2. 
$$
Note that these matrices can be equivalently defined as conditionally negative matrices with zero diagonal, see~\cite{scholkopf2002learning}.

\begin{proposition}\label{prop:mmd}
One has $\Phi_A^\infty : \Kk_m \rightarrow \Kk_n$ where 
$$
\Phi_A^\infty(\C) \eqdef \lim_{\eps\to\infty}\Phi_\A^\eps(\C) \!=\! ( -\ufrac{2}\dotp{\C(\a_k-\a_\ell)}{\a_k-\a_\ell} )_\kl.$$
\end{proposition}

This property shows that in the large $\eps$ limits $\Phi_A^\infty$ is actually a linear map which computes Maximum Mean Discrepancies~\cite{gretton2012kernel} (a.k.a. Euclidean distances between probability distributions).

\myparagraph{Connection with PCA} For the sake of simplicity in the exposition, let us now assume $A=B^\top$.
In this case, \eqref{eq-wass-sing-problem} is a classical linear singular vectors problem. While in general, ensuring existence of positive singular vectors is non-trivial, the following proposition, proved in Appendix~\ref{proof-prop-pca}, shows that this is the case for $\Phi_A^\infty$.

\begin{proposition}[Connection with PCA]
\label{prop-pca}
    Let us denote $\tilde\A \eqdef \A- \A\ones_m \ones_m^\top/m$ the centered matrix.
    For any pair $(u,v)$ of singular vectors of $\tilde\A$ with singular value $\lambda$, 
    $$
    \C=( (v_k - v_\ell)^2)_\kl\in\Kk_m, 
    \;
    \D=( (u_i - u_j)^2)_\ij\in\Kk_n
    $$
    are singular vectors of $(\Phi_\A^\infty,\Phi_\B^\infty)$,
    with singular value $2\lambda^2$.
\end{proposition}

This proposition shows that for $\eps=+\infty$ a set of positive singular vectors is obtained as simply squared Euclidean distances over 1-D principal component embeddings of the data. Entropic regularization thus draws a link between our novel set of OT-based metric learning techniques and classical dimensionality reduction methods. This frames Sinkhorn singular vectors as a well-posed problem regardless of the value of $\eps$.

\section{Metric Learning for Single-Cell Genomics}
\label{sec-genomics}

\begin{figure}[h]
    \centering
    \includegraphics[width=.45\textwidth]{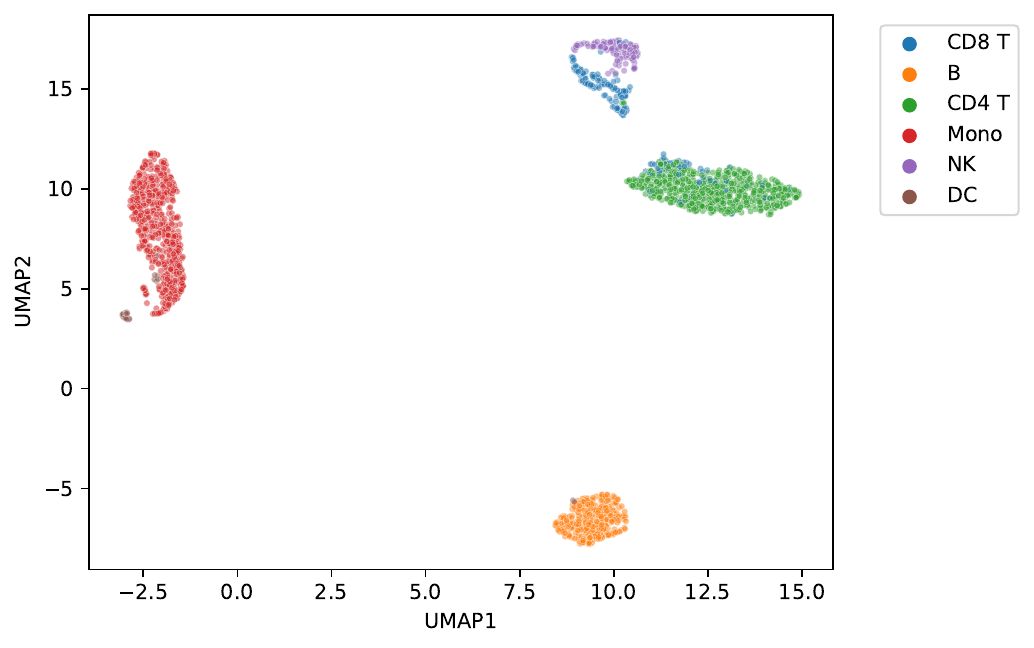}
    \caption{UMAP projection of the cells of a scRNA-seq dataset using the singular vector $\D$, with cells colored by cell type.}
    \label{fig:results_rna_cell}
\end{figure}

\myparagraph{scRNA-seq data} Single-cell RNA sequencing (\textit{scRNA-seq}) is a high-throughput sequencing technology enabling the measurement of gene expression levels at single-cell resolution \cite{stegle2015computational}. The analysis of scRNA-seq data has offered unprecedented insights in cellular heterogeneity and disease mechanisms \cite{tanay2017scaling,yuan2017challenges}.
scRNA-seq data can be represented as a matrix of integer expression levels with cells on rows and genes on columns. One of the main uses of scRNA-seq is to identify cell populations through clustering or visualization. But these tasks rely on some notion of distance between cells. The most popular clustering and visualization tools, in particular Scanpy~\cite{wolf2018scanpy} and Seurat~\cite{stuart2019comprehensive}, rely on Euclidean distances on PCA embeddings of cells. Embeddings can also be provided by deep learning models like scVI~\cite{gayoso2022python}. A good metric on the space of genes is also important because the phenotype of a cell is determined by the joint activity of all its expressed genes.

\myparagraph{Optimal Transport distances between cells} In order to take advantage of the biological relationships between genes, OT distances between cells have recently been proposed. The Gene Mover Distance \cite{bellazzi2021gene} is defined similarly to the Word Mover Distance~\cite{kusner2015word}: the authors use as a ground cost the Euclidean distance between precomputed Gene2Vec~\cite{du2019gene2vec} embeddings. \cite{huizing2021optimal} use a Sinkhorn divergence with a cosine distance between genes (i.e. vectors of cells) as a ground cost. In the present paper we compute OT distances using the Python package POT~\cite{flamary2017pot}.

\myparagraph{Dataset} A commonly analyzed scRNA-seq dataset is the ``PBMC 3k" dataset produced by 10X Genomics, obtained through the function \texttt{pbmc3k} of Scanpy~\cite{wolf2018scanpy}. Details on preprocessing and cell type annotation are given in Appendix~\ref{supp-data-processing}. The processed dataset contains $m=1030$ genes and $n=2043$ cells, each belonging to one of 6 immune cell types: `B cell', `Natural Killer', `CD4+ T cell', `CD8+ T cell', `Dendritic cell' and `Monocyte'. The cell populations are heavily unbalanced. In addition, for each cell type we consider the set of canonical marker genes given by Azimuth~\cite{hao2021integrated}, i.e. genes whose expression is characteristic of a certain cell type.

\begin{figure*}[h]
    \centering
    \includegraphics[width=.43\textwidth]{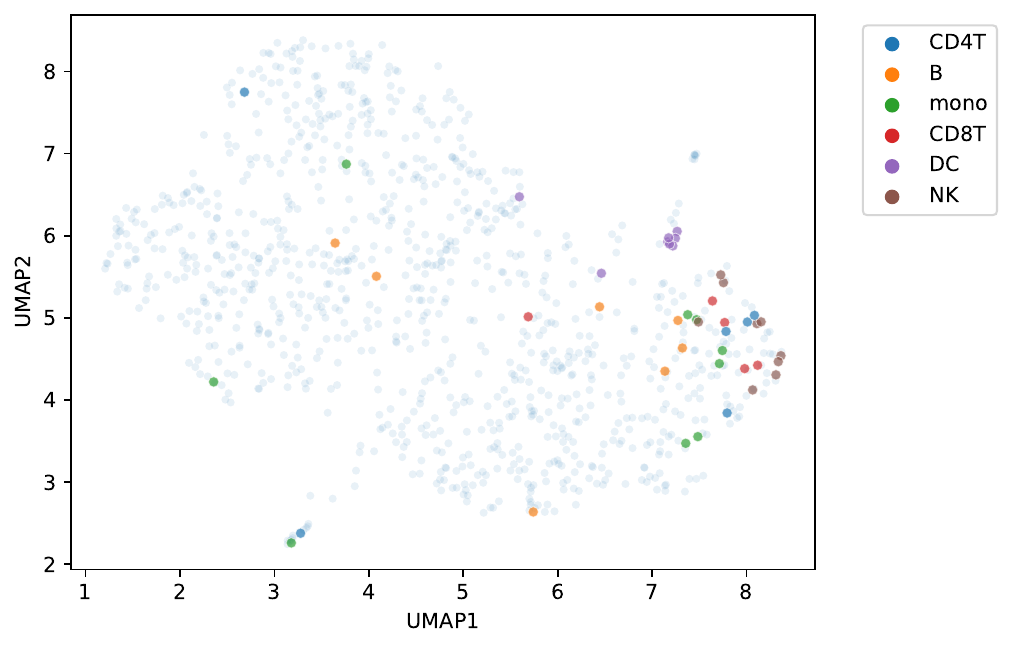}
    \includegraphics[width=.43\textwidth]{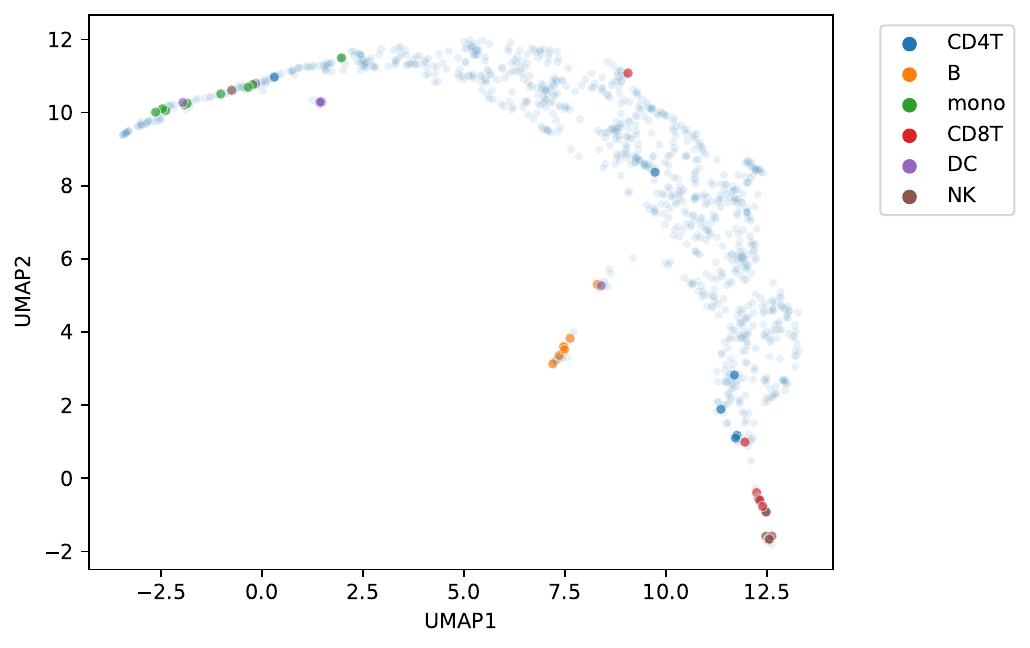}
    \caption{2-D UMAP projection of marker genes, using the computed distances. Marker genes are colored by associated cell type, and other genes are faded out. Left, the Euclidean distance on Gene2Vec~\cite{du2019gene2vec} embeddings. Right, the singular vector $\C$.}
    \label{fig:results_rna_gene}
\end{figure*}

\myparagraph{Evaluation} We use the annotation on cells (resp. on marker genes) to evaluate the quality of distances between cells (resp. between marker genes). We report in Table~\ref{tab:results_rna_cell} and Table~\ref{tab:results_rna_gene} the Average Silhouette Width (ASW), computed using the function \texttt{silhouette\_score} of Scikit-learn~\cite{scikit-learn}. In addition, we visualize both of these distances using a 2-D UMAP projection~\cite{mcinnes2018umap}. We compare
(i) Euclidean distances on PCA embeddings
(ii) Euclidean distances on Kernel PCA embeddings using Scikit-learn's implementation with \texttt{kernel='rbf'}
(iii) Euclidean distances on scVI~\cite{gayoso2022python} embeddings using default values
(iv) Gene Mover Distance
(v) Sinkhorn divergence ($\eps=.1$) with a cosine distance between genes as a ground cost 
(vi) Wasserstein Singular Vectors ($\tau=0.001, \eps=0.1$), approached by 15 power iterations, which we found in this case to lead to better results than the stochastic power iterations. Note that for large regularisation $\eps$, as shown in Section~\ref{sec-entropic}, the Wasserstein Singular Vectors are themselves (squared) Euclidean distances on PCA embeddings.

\myparagraph{Results} The results in Table~\ref{tab:results_rna_cell} and Table~\ref{tab:results_rna_gene} suggest that our method improves over all considered baselines. Figure~\ref{fig:results_rna_cell} and Figure~\ref{fig:results_rna_gene} shows the UMAP projection of the cells and the genes in the dataset. The Wassersein Singular Vectors clearly outperform the other metrics in terms of Average Silhouette Widths, both in the context of cells and of genes. Interestingly, in the case of marker genes we outperform the Euclidean distance on Gene2Vec embeddings, which are meant to contain ``semantic" information about genes. These scores are also validated by the UMAP projection, where cells and marker genes cluster according to cell type. These results motivate further research in the biological implications of Wasserstein Singular Vectors.
Let us highlight that we compute distances between cells or genes, but that we do not produce embeddings like PCA or scVI. In addition, scVI can handle complex tasks like the removal of unwanted sources of variation~\cite{gayoso2022python} which we do not consider in this article.

\begin{table}[h]
    \centering
    \caption{Average Silhouette Width for cells}
    ~\\
    \begin{tabular}{ll}
        \hline
        Method & ASW\\
        \hline
        PCA / $\ell^2$ & 0.238\\
        Kernel PCA / $\ell^2$ & 0.241\\
        scVI embedding / $\ell^2$ & 0.168\\
        Sinkhorn & 0.003\\
        Gene Mover Distance & 0.066\\
        WSV (ours) & \textbf{0.348}\\
        
    \end{tabular}
    \label{tab:results_rna_cell}
\end{table}

\begin{table}[h]
    \centering
    \caption{Average Silhouette Width for marker genes}
    ~\\
    \begin{tabular}{ccc}
        \hline
        $\ell^2$ & Gene2Vec / $\ell^2$ & WSV (ours) \\
        \hline
        -0.005 & 0.0186 & \textbf{0.136}
    \end{tabular}
    \label{tab:results_rna_gene}
\end{table}
\vspace{-.3em}

\section{Conclusion and Perspectives}

Wasserstein Singular Vectors define a pair of ``intrinsic'' ground metrics associated to a given dataset. This elegantly solves the problem of unsupervised ground metric learning without resorting to ad hoc embeddings. 
Numerical results on single-cell RNA sequencing suggest that these metrics encode salient geometric structures of the data. 
This opens several avenues for future works, in particular an in-depth theoretical analysis when $\tau=0$ and $\eps>0$. 
Our method can be extended to unbalanced optimal transport~\cite{liero2015optimal,chizat2018unbalanced}, which has proved useful to increase the robustness of the metric for the analysis of biological sequencing datasets~\cite{schiebinger2019optimal}. Lastly, our initial results regarding stochastic approximation of Wasserstein Singular Vectors would greatly benefit from further developments enabling better convergence rates.

\section*{Acknowledgements}
We thank Stéphane Gaubert for very useful advice on non-linear Perron-Frobenius theory.

This work was performed using HPC resources from GENCI-IDRIS [Grant 2021-AD011012285]. The project leading to this publication has received funding from the Agence Nationale de la Recherche (ANR) project scMOmix and Sanofi iTech Awards. The work of G. Peyré is supported by the European Research Council (ERC project NORIA) and by the French government under management of Agence Nationale de la Recherche as part of the ``Investissements d’avenir'' program, reference ANR19-P3IA-0001 (PRAIRIE 3IA Institute).

\bibliography{mybiblio}
\bibliographystyle{icml2022}

\clearpage
\pagebreak
\appendix
\onecolumn

\section{Computation Times and Numerical Resources}
\label{supp-resources}

CPU computations were performed on a Dell Latitude 5420 with an 8 core 11th Gen Intel(R) Core(TM) i7-1165G7 @ 2.80GHz CPU. GPU computations were performed on Nvidia V100 SXM2 32 Go GPUs.

50 pairs of Wasserstein power iterations for the synthetic datasets in Section~\ref{numerical-translated} run in about three minutes on CPU.

15 pairs of Sinkhorn ($\eps=0.1$) power iterations for the single-cell dataset in Section~\ref{sec-genomics} run in about 1h50mn on GPU.

\section{Proof of Theorem~\ref{thm-existence} (Existence of Singular Vectors)}
\label{proof-existence}
We consider the set $$\Kk_n^\rho \eqdef \{\D\in\Dd_n\st \norm{\D}_\infty=1, \forall i\neq j, \D_\ij \geq \rho\}$$

Let $(\C, \D)\in\Kk_m^\rho\x\Kk_n^\rho$. A classical result states that
$$\dfrac{\rho}{2} \norm{\a - \a'}_1 \leq \Wass_\C(\a, \a') \leq \ufrac{2}\norm{\a - \a'}_1$$

Thus with $(\C', \D') = \left(\dfrac{\Phi_\B(\D)}{\norm{\Phi_\B(\D)}_\infty}, \dfrac{\Phi_\A(\C)}{\norm{\Phi_\A(\C)}_\infty}\right)$,
$$
    \frac{\rho}{2} \norm{\b_k - \b_\ell}_1 + \tau R(\b_k - \b_\ell) \leq \Phi_\B(\D)_\kl \leq \ufrac{2}\norm{\b_k - \b_\ell}_1 + \tau R(\b_k - \b_\ell)
$$
and
$$
    \frac{\rho}{2} \norm{\a_i - \a_j}_1 + \tau R(\a_i - \a_j) \leq \Phi_\A(\C)_\ij \leq \ufrac{2}\norm{\a_i - \a_j}_1 + \tau R(\a_i - \a_j).
$$
Equivalence of norms implies $k_R^-\norm{\cdot}_1 \leq R(\cdot)\leq k_R^+\norm{\cdot}_1$. With  $\gamma_A\eqdef \dfrac{\min_{i\neq j} \norm{\a_i - \a_j}_1}{\max_{i\neq j} \norm{\a_i - \a_j}_1}$ and $\gamma_B\eqdef \dfrac{\min_{k\neq \ell} \norm{\b_k - \b_\ell}_1}{\max_{k\neq l} \norm{\b_k - \b_\ell}_1}$,
$$
    \forall k\neq \ell, \C'_\kl \geq \dfrac{\rho + 2\tau k_R^-}{1 + 2\tau k_R^+}\gamma_\B \mathand \forall i\neq j, \D'_\ij \geq \dfrac{\rho + 2\tau k_R^-}{1 + 2\tau k_R^+}\gamma_\A.
$$
This shows that for
$$
    0 < \rho \leq \min\left(\dfrac{2\gamma_\A\tau k_R^-}{2\tau k_R^+ + 1 - \gamma_\A}, \dfrac{2\gamma_\B\tau k_R^-}{2\tau k_R^+ + 1 - \gamma_\B}\right)
$$
one has $(\C', \D')\in\Kk_m^\rho\x\Kk_n^\rho$. So for such $\rho$, the function $(\C', \D') \mapsto \left(\dfrac{\Phi_\B(\D)}{\norm{\Phi_\B(\D)}_\infty}, \dfrac{\Phi_\A(\C)}{\norm{\Phi_\A(\C)}_\infty}\right)$ is a continuous map from the locally contractible compact $\Kk_m^\rho\x\Kk_n^\rho$ to itself 
so using the Brouwer theorem, it has a fixed point, which is a pair of singular vectors of ($\Phi_\A$, $\Phi_\B$).

\section{Proof of Proposition~\ref{prop:contractlarge} (Uniqueness of Singular Vectors for Large $\tau$)}
\label{unique-large-tau}

Denoting $s = 1/\tau$, we consider the map
$$
    \Psi_\A(\C) \eqdef \frac{ 
        \norm{\C}_\infty U + s W_A(\C)  
    }{
        \norm{ \norm{\C}_\infty U + s W_A(\C) }_{\infty}
    }
$$
on the set of $\norm{\C}_\infty = 1$, where $U \eqdef (R(a_i-a_j))_{i,j}$ is constant and $W_A(\C) \eqdef (\Wass_\C(\a_i,\a_j))_{i,j}$. 
Since $W_A$ is 1-Lipschitz, and $\norm{\C}_\infty=1$, one needs to study the contractance of 
$$
    W \mapsto \frac{ 
         U + s W  
    }{
        \norm{ U + s W  }_\infty
    }.
$$
One can explicitly compute the derivative of this map, which is $O(s)$, so that $\Psi_\A$ is itself contractant for $s$ small enough. This shows that $(\C,\D) \mapsto (\Psi_\B(\D), \Psi_\A(\C))$ is also contractant, which implies uniqueness of the singular vector and linear convergence for $\norm{\cdot}_\infty$ of the power iterations.

\section{Additional Figure for the Numerical Illustration of Section~\ref{numerical-translated}}
\label{supp-numeric-translated}

\begin{figure}[h]
    \centering
    \includegraphics[width=\textwidth]{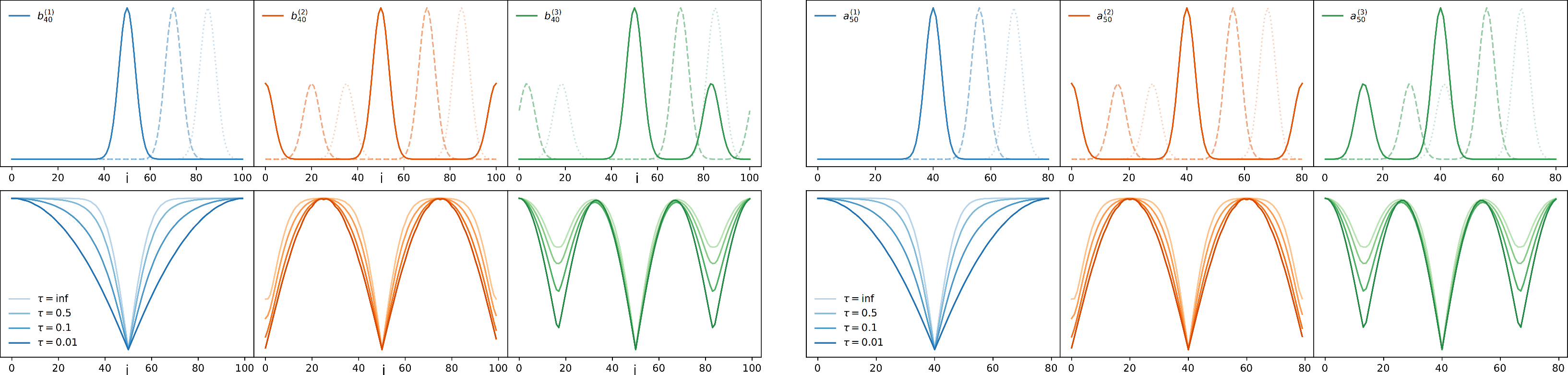}
    \caption{(left, top) Translated histograms template defining $\B_p$ (left, bottom) One element of the singular vector $\C_p$ (right, top) Translated histograms template defining $\A_p$ (right, bottom) One element of the singular vector $\D_p$.}
    \label{fig:supp_synth}
\end{figure}

Figure~\ref{fig:supp_synth} shows that as predicted, the singular vectors $\C_p$ and $\D_p$ are identical up to rescaling.

\section{Proof of Convergence for the Stochastic Power Iterations of Section~\ref{thm-stochastic}}
\label{proof-thm-stochastic}
\begin{proof}
We follow steps similar to the proof of \cite{nemirovski2009robust} for projected stochastic gradient descent. The theorem supposes $\tau$ large enough, so we can consider a pair of unique Wasserstein Singular Vectors $(\C^\star, \D^\star)$. We study the quantity $\EE[\ell_{t}]$ where $\ell_{t}\eqdef\norm{\C_{t} - \C^\star}_2^2$. The proof for $\norm{\D_{t} - \D^\star}_2^2$ is identical. We consider a constant scaling factor $\tilde\lambda$. Let us start by defining
$$
    (\TT_{t})_\kl = \begin{cases}
        (\C_{t})_\kl - \Phi_\B(\D_{t})_\kl/\tilde\lambda \textrm{ if } (\kl) = (k_t,\ell_t)\\
        0 \textrm{ otherwise}\\
    \end{cases}$$ and $\TT^\star \eqdef p(\C^\star - \Phi_\B(\D^\star)/\tilde\lambda)
$
where $p=\ufrac{mn}$ is the probability for an element to be updated.

By definition of the power iterations, $\C_{t+1} = \Pi(\C_{t} - \alpha_{t}\TT_{t})$.

By definition of the Wasserstein Singular Vectors, $\C^\star = \Pi(\C^\star - \alpha_{t}\TT^\star)$.

Thus, $$\norm{\C_{t+1} - \C^\star}_2^2 = \norm{\Pi(\C_{t} - \alpha_{t}\TT_{t}) - \Pi(\C^\star - \alpha_{t}\TT^\star)}_2^2.$$

The value of $\tilde\lambda$ proposed in the theorem ensures that $\norm{\C_{t} - \alpha_{t}\TT_{t}}_\infty \geq 1$ and $\norm{\C^\star - \alpha_{t}\TT^\star}_\infty \geq 1$.

The theorem of projection on a convex (the unit sphere for the norm $\norm{\cdot}_\infty$) then ensures that
$$\norm{\C_{t+1} - \C^\star}_2^2 \leq \norm{(\C_{t} - \C^\star) - \alpha_{t}(\TT_{t} - \TT^\star)}_2^2.$$
Decomposing the squared norm, we get
\begin{align*}
    &\norm{(\C_{t} - \C^\star) - \alpha_{t}(\TT_{t} - \TT^\star)}_2^2 =\\
    \ell_{t} - 2 & \alpha_{t}\dotp{\C_{t} - \C^\star}{\TT_{t} - \TT^\star} + \alpha_{t}^2\norm{\TT_{t} - \TT^\star}_2^2.
\end{align*}
The middle term can be simplified when taking its expectation:
\begin{align*}
    &\EE_{t}[\dotp{\C_{t} - \C^\star}{\TT_{t} - \TT^\star}] =\\ p\ell_{t} - \frac{p}{\tilde\lambda}\dotp{\C_{t} &  - \C^\star}{ \Wass(\D_{t}) - \Wass(\D^\star)} \geq (1 - L/\tilde\lambda)p\ell_{t}, 
\end{align*}
for some constant $L$, since $\Wass$ is 1-Lipschitz with regards to the infinite norm as proved earlier. The last term can be bounded as well:
$$\norm{\TT_{t} - \TT^\star}_2^2 \leq \norm{\TT_{t}}_2^2 + \norm{\TT^\star}_2^2 \leq 2p\max(1, \tau\norm{R_\B}_\infty/\tilde\lambda).$$

Calling that term $M$, we have finally $$\EE_{t}[\ell_{t+1}] \leq \left(1 - 2\alpha_{t}p\left(1 - L/\tilde\lambda\right)\right)\x \ell_{t} + \alpha_{t}^2M.$$

In the next steps we assume $\tau$ big enough for $p\left(1 - L/\tilde\lambda\right)$ to be positive and name the quantity $Q$.

Let us note that an overly pessimistic upper bound for $L$ is $m$, which would require a very large value of $\tau$. However, this upper-bound of $L$ is obtained by juggling between different norms. In practice, the algorithm converges for arbitrarily small values of $\tau$. This suggests a much smaller constant, that does not depend on the data's dimensionality.

Taking the expectation over all times $t$,
$$\EE[\ell_{t+1}] \leq \left(1 - 2\alpha_{t}Q\right)\x \EE[\ell_{t}] + \alpha_{t}^2M.$$

Reformulating,
$$ 2\alpha_{t} Q \EE[\ell_{t}] \leq \EE[\ell_{t}] - \EE[\ell_{t+1}] + \alpha_{t}^2M.$$

Summing along $t=1...T$,
$$ \min_{t=1...T}\EE[\ell_{t}] \leq \left(\sum_{t=1}^T \alpha_{t}\right)^{-1}\left(\dfrac{\ell_1}{2Q} + \dfrac{M}{2Q}\sum_{t=1}^T\alpha_{t}^2\right).$$

For $\alpha_t=1/\sqrt{t}$, we thus have classically a convergence rate of $\Oo(\log(t)/\sqrt{t})$
\end{proof}

\section{Sinkhorn Algorithm}
\label{supp-sinkhorn}

The Sinkhorn cost can be computed by the dual formula
$$
	W_{\C}^\eps(a_i,a_j) = \eps\left( \dotp{\log(u)}{a_i}+\dotp{\log(v)}{a_j} - \dotp{K v}{u} \right),
$$
where $K\eqdef\exp(-\frac{\C}{\eps\norm{\C}_\infty})$ and $(u,v)$ are obtained by iterating the following Sinkhorn fixed point
$$
	u \leftarrow \frac{a_i}{K^\top v}
	\mathand
	v \leftarrow \frac{a_j}{K u}.
$$
This allows one to compute with precision $\eps$ the $n^2$ entries of $\Phi_A^\eps(\C)$ in $\Oo( (m n)^2/\eps^2 )$ operations~\cite{altschuler2017near}, using a parallelizable algorithm that is well suited for GPU computations.

\section{Proof of Proposition~\ref{prop-pca} (Connection with PCA)}
\label{proof-prop-pca}
\begin{proof}
We define the operator mapping correlation kernels to Euclidean distance
$$
    \Delta(K) \eqdef -(K + K^\top) + \diag(K)\ones_n^T + \ones_n\diag(K)^\top.
$$
One has the convenient formula
$$
    \Phi_\A^\infty(\C) = -\Delta(\A^\top \C~ \A).
$$
Let the centering operator $J \eqdef \Id_n-\ones_{n \times n}/n$, which satisfies $J^2=J$ and $\ker(J)=\mathrm{Span}(\ones_n)$.

Let $u\in\RR^n$ and $v\in\RR^m$ be a pair of singular vectors, so that there exists $\lambda\in\RR$ such that 
$$
    A J u = \lambda v
    \quad\text{and}\quad
    J A^\top v = \lambda u.
$$
By linearity and using the fact that 
$$
    \ker(\Phi_\A^\infty) = \{C \st C = -C^\top\}\cup\{\a\ones_n^\top + \ones_n \b^\top\}
$$
one has 
\begin{align*}
    \Phi_\A^\infty(\Delta(vv^\top)) &= -2\Phi_\A^\infty(vv^\top)\\
    &= 2\Delta(A^\top vv^\top A)\\ &= 2\Delta(JA^\top vv^\top AJ)\\ &= 2\lambda^2\Delta(uu^\top),
\end{align*}

where we used the fact that $\Delta(JKJ) = \Delta(K)$.
The same reasoning for $\Phi_\B^\infty$ yields the advertised result.
\end{proof}

\section{Details on Data Processing}
\label{supp-data-processing}

We recovered the `pbmc3k' dataset using the function \texttt{pbmc3k} of Scanpy~\cite{wolf2018scanpy}. Cell types were annotated using the Azimuth~\cite{hao2021integrated} web tool, which projects it onto large-scale reference atlases. We removed the cluster `other T cells' and cells for which the annotation was less than 90\% confident. Cells were selected using a standard quality filtering pipeline. The data was CPM-normalized, log1p-transformed, and then the 1000 most varying genes were selected. To those genes we added the canonical markers given in the documentation of Azimuth~\cite{hao2021integrated}.

\end{document}